\documentclass[sigconf]{acmart}
\usepackage{xurl}
\usepackage{xparse}
\AtBeginDocument{%
  \providecommand\BibTeX{{%
    \normalfont B\kern-0.5em{\scshape i\kern-0.25em b}\kern-0.8em\TeX}}}

\setcopyright{rightsretained}
\copyrightyear{2023}

\acmConference[HRCI Workshop '23]{Workshop on Human-Robot Conversational Interaction}{March 13th,
  2023}{Stockholm, SE}
\begin{document}

\title{Sequential annotations for naturally-occurring HRI: first insights}


\author{Lucien Tisserand}
\affiliation{%
  \institution{UMR 5191 ICAR, CNRS, Univ Lyon, ENS Lyon}
  \city{Lyon}
  \country{France}}
\email{lucien.tisserand@ens-lyon.fr}

\author{Frédéric Armetta}
\affiliation{%
  \institution{Univ Lyon, UCBL, CNRS, INSA Lyon, LIRIS, UMR5205, F-69622}
  \city{Villeurbanne}
  \country{France}}
\email{frederic.armetta@liris.cnrs.fr}

\author{Heike Baldauf-Quilliatre}
\affiliation{%
  \institution{UMR 5191 ICAR, CNRS, Univ Lyon, ENS Lyon}
  \city{Lyon}
  \country{France}}
\email{heike.baldaufquilliatre@ens-lyon.fr}

\author{Antoine Bouquin}
\affiliation{%
  \institution{Univ Lyon, UCBL, CNRS, INSA Lyon, LIRIS, UMR5205, F-69622}
  \city{Villeurbanne}
  \country{France}}
\email{antoine.bouquin@liris.cnrs.fr}

\author{Salima Hassas}
\affiliation{%
  \institution{Univ Lyon, UCBL, CNRS, INSA Lyon, LIRIS, UMR5205, F-69622}
  \city{Villeurbanne}
  \country{France}}
\email{salima.hassas@liris.cnrs.fr}

\author{Mathieu Lefort}
\affiliation{%
  \institution{Univ Lyon, UCBL, CNRS, INSA Lyon, LIRIS, UMR5205, F-69622}
  \city{Villeurbanne}
  \country{France}}
\email{mathieu.lefort@liris.cnrs.fr}

\renewcommand{\shortauthors}{Tisserand et al.}

\begin{abstract}
    We explain the methodology we developed for improving the interactions accomplished by an embedded conversational agent, drawing from Conversation Analytic sequential and multimodal analysis. The use case is a Pepper robot that is expected to inform and orient users in a library. In order to propose and learn better interactive schema, we are creating a corpus of naturally-occurring interactions that will be made available to the community. To do so, we propose an annotation practice based on some theoretical underpinnings about the use of language and multimodal resources in human-robot interaction.
\end{abstract}



\received{12 Feb. 2023}
\received[accepted]{1st March 2023}
\received[revised]{3rd March 2023}

\maketitle

\section{Introduction: project goals}

This methodological paper draws from an ongoing project called Peppermint\footnote{Full title "Interacting with Pepper: Mutual Learning of Turn-taking Practices in HRI" (2021-2024). Project website: \url{https://peppermint.projet.liris.cnrs.fr/}}.
As part of this project, \emph{Conversation Analysis} researchers and \emph{Artificial Intelligence} researchers team up in a collaborative effort to improve interactions in the wild with an autonomous
Pepper\footnote{Manufactured by Aldebaran. Please visit \url{https://www.aldebaran.com/fr/pepper}} robot as regard with turn-taking and action recognition with the perspective of sequential organization.
The robot's use case is offering information and orientation services in a university library in France.

We follow an inductive, step-by-step approach that rely on the production and analysis of naturally-occurring data.
In short, we first created an \textit{ad hoc} autonomous conversational system as a state machine. 
This first software allowed us to video-record naturally-occurring data (see \ref{naturally}) of \emph{human-robot interactions} (\textit{HRI}) for empirical, inductive findings (of which are \textit{HRI} specifics).
We are now structuring this corpus and annotating it with regard to a core principle of human interaction: \emph{sequence and sequential organization}.
This dataset will be used to improve conversational \textit{HRI} by using machine learning / NLU methods.

In this paper, we first explain what it means to consider sequential organization as a temporal, continuous achievement of mutual understanding, and its relevance for having a conversational system to respond appropriately and timely (\ref{theory}).
We then explain how a heterogeneous dataset of naturally-occurring HRI can be systematically managed through a labelling scheme (\ref{corpus}).
We finally sketch an annotation syntax addressing sequential organization (\ref{annotation}) before discussing its potential (\ref{discussion}).

\section{Theoretical Underpinnings}
\label{theory}
Within this section, we explain how the \emph{Conversation Analytic} (\textit{CA}) approach to human interaction provides new insights on the analysis and annotation of a dataset that account for the sequential organization of multimodal HRI. Especially, we focus on the dynamics of normative expectations (vs. predictions) and interpretative feedbacks that allow a completely unique and unpredictable interaction to be controlled on a turn-by-turn basis.

From a \emph{CA} perspective, the analysis of talk and gestures in interaction is above all the analysis of how talk (e.g. lexical choice, intonation...) and other resources (body position, gaze, gestures...) are in fact designed to be used in interaction~\cite{mondada2014multimodal}. This vision is opposed to intrinsically meaningful conducts that would simply be adapted to an interaction setting. Growing on \emph{ethnomethodological} roots, \textit{CA} shows how the mutual understanding is "an operation rather than a common intersection of overlapping sets" as Garfinkel puts it~\cite[p.~30]{garfinkel1967}. As we will see, such process is achieved by the mean of sequential organization through \textit{turn-taking} and it implies that participants formulate turns for accomplishing contextually relevant actions.

\subsection{Mutual understanding in human-human interaction: bottom-up and top-down}

\emph{CA} has been partly founded on the investigation of noticeable regularities with regard to the accomplishment of turn-taking practices such as the transitions without gaps and overlaps despite the fact that turns have various durations~\cite{sacksetal1974}. The management of these regularities led to the idea that turns are composed of units (\emph{Turn-Constructional Units}) processed by a commonly shared \textit{turn-taking system} (\textit{TTS}) which rules have been described.
This system has sometimes only been seen as a finely tuned \textit{bottom-up} mechanics~\cite{sharrock2000}, and some streams of research focused on such units~\cite{fordthompson1996,fordetal2015,selting2000}, how these were implemented by an analyzable signal and some explored their computational processing~\cite{skantze2020}.

But while it may appear relevant to study how turn-ending could be identified in order to handle turn-taking (such as~\cite{skantze2020}), we consider the other route, a more \textit{top-down} approach: interactants do not take turns for the sake of taking turns, they do so in order to create a delimited and purposeful context for future interpretations of actions and intents that will be implemented by talk and bodily conducts.
By doing so, they accomplish collaborative activities while continuously ensuring that what had to be interpreted was actually interpreted as such.
Speakers make use of a \textit{sequential} approach to interpretation: a next speaker's conduct is always interpreted within a slot temporally projected by a prior action even if the next action is ultimately interpreted as a completely unexpected next move.
That functioning leads to the fact that next speakers do display such departures from projected next turns. For example, in the imaginary case below, it is indicated with the turn-beginning \texttt{"well,hum"} followed by an extended account:

\small
\texttt{A: "Hello, can I help you ?"}

\texttt{B: "Well, hum, I'm just waiting for my friend."}
\normalsize

We might think of a commonly shared inventory of such contextual practices with the notion of \textit{adjacency pairs} that draws on the idea that sequences of actions are culturally typified as normative pairs (greeting-greeting, offer-acceptance/reject...)~\cite{schegloff2007}. Thanks to the turn-taking organization, different speakers participate alternatively to the first pair part or second pair part. They do not follow a set of rules or instructions that will determine their conducts, they refer to this norm in order to ease the action ascription of turns~\cite{levinson2013,button1990,buttonsharrock1995}. Although this "repertoire" vision is limited when it comes to grasp the complexity of human-human interaction~\cite{enfieldsidnell2022}, it appears adapted to the human "simplistic" approach to service encounter HRI (see \ref{discussion}).

If some verbal and multimodal resources participate to the \textit{bottom-up} recognition of such actions (e.g. a Wh-questions projecting types of responses at turn-beginning~\cite{deppermann2012}), sequence organization and adjacency pairs are crucial \textit{top-down} resources for the recognition of actions in a delimited context~\cite{levinson2013} and thus the recognition of turn completions~\cite{levinsontorreira2015}.
From that perspective, that also means that the "right" interpretation of a turn is the understanding of what can be produced next for all practical purpose (\textit{versus} the semantic management and selection of all interpretable actions and meanings of a verbal turn).

\subsection{Temporal, turn-by-turn increments}
\label{temporally}
As explained above, humans make use of sequentiality to incrementally secure their interpretation of what they are expected to do next, and arguments point toward universals with regard to such infrastructure~\cite{kendricketal2020}.
Just to give a glimpse of all intricate sequences that implies, participants may recognize and accomplish the answers that are expected~\cite{stiversrobinson2006}, but they may also initiate \textit{repairs}~\cite{hayashietal2013} projecting \textit{reformulations} by the previous speaker, they may reformulate themselves what they understood~\cite{deppermann2011} or produce feedback during the turn~\cite{goodwin1986}.
Moreover, these methods may be used at different places as regard with the \textit{adjacency pair} organization: before (e.g. "what I wonder is") , in-between (e.g. "what do you wanna know?") or after ("okay great").
Some previous turns or whole sequences may be reformulated, expanded, but also normative expectations may just be abandoned.
 
Thus, the approach to the modelization of the temporal trajectory of an interaction might imply the suspension of the immediacy of second pair parts or following sequences. This turn-by-turn temporally incremented display of successive interpretations is at the heart of the mutual understanding process in interaction~\cite{mondada2011}. As Levinson~\cite[p.~47]{levinson2006} recalls us, the representation of successive turns of a human-human interaction is then less a linear representation like \texttt{[A1->B1->A2->B2]} than some kinds of stacking structures like \texttt{[A1->B2->A2->B1]}, where letters are the interactants, and numbers distinguishing sequence types.
These are the structures we aim at investigating and annotating.

\subsection{What about artificial conversational agents ?}

If we apply this perspective to the design of an autonomous conversational agent, a quite reluctant implication is that no word-based treatment of the human input is sufficient to ascribe the human turn to actions (\textit{bottom-up dead} end). Moreover, the sequence organization being not a set of instructions but a set of conventions, every next move is virtually possible, and no pre-drawn scheme of action can be hard-coded (\textit{top-down} dead end).

A more attractive perspective is to consider the fact that humans make use of turns and norms as a way to produce more flexible and negotiable interpretations and that speakers leave cues that make these practices recognizable like the "well hum" above.
The criterion for a successful next turn is the formulation of a possible next that projects further sequences, which means that there is more than only one "good answer" produced by the machine. What can then be investigated is the set of procedures that humans rely on in order to make sense of every next turn.

When it comes to the design or analysis of conversational agents, several researchers explore the benefits of the incremental dimension of interaction for securing a face-to-face encounter with an autonomous and responsive system, like Fischer and Sikveland~\cite{fischersikveland2019} in the case of what Stivers and Robinson have identified as \textit{progressivity}~\cite{stiversrobinson2006}, or Julian Hough for self-repair practices~\cite{hough2015}. Housley and colleagues~\cite{housleyetal2019} advocated for collaborative attempts to apply CA to AI in the case of big interactional data with a sequence-of-action oriented approach (\textit{vs.} linguistic features or emotional cues). The attempts we present in this methodological and reflexive paper can be read as one way to go in that direction.

In fact, considering the sequential infrastructure (as a set of possible next, preferred next, insertions, expansions, projections of series of sequences...) for computing, as CA already established it, is a proposition that goes back more than thirty years ago, when Gilbert, Wooffitt and Fraser\cite{gilbert_organising_1990} addressed the fact that the sequence analysis drawing on adjacency pairs could be subject to formalisation, although these would not explain all the contextual cues that participate to mutual understanding agreement. 
This initiative was duly demotivated~\cite{button1990,buttonsharrock1995} by arguments that we mentioned and that we also agree with: the contingent and contextual character of interpretations, the non-scriptable character of interactions, the conventional (vs. instructional) character of sequence organization. 

However, back in the days, what was discussed was the possibility to hard-code such grammar rules for the management of turns in interaction and for sentence/action recognition, as a deductive approach. Given the progress that have been made into the automatic discovery of statistical/probabilistic rules from annotated datasets (both in Natural Language Processing and Understanding), even with a small number of tokens (in the case of few shots learning), we advocate that it worth trying to rely on complex sequential annotations and adequate algorithms in order to provide a conversational agent with a statistically-oriented sequence management for all practical purpose. Moreover as conversational user interfaces are now ubiquitous in various societies, people display an alignment with such functioning (see~\cite{Porcheron2018} and \ref{discussion}).

\section{Corpus construction}
\label{corpus}
Within this section, we present how we acquired the data in order to make sure we would obtain naturally-occurring data that account for the complex sequential and multi-party dimensions of interactions. By essence, such a corpus is heterogeneous, that is why we explain how we manage this through a labelling scheme~(\ref{shortclips}).

\subsection{Use case}
The dialog proposed by Pepper was based on a state machine (see appendix \ref{appendix-documentation} for the details) where the transitions rely on the detection of some specific words. We used the manufacturer's built-in
APIs\footnote{Mainly the QiChatbot API using QiChat script language. Please visit \url{https://qisdk.softbankrobotics.com/sdk/doc/pepper-sdk/ch4_api/conversation/qichat/qichat_index.html}}
 for word detection and prompt-to-answer rules.
In order to anticipate user questions, we asked what were the most simple and recurrent requests that the library users were asking to the reception desk agents (location of toilets, how to connect to the wi-fi...).
We asked them what they would like to see accomplished by a robot, so that it could be seen as a alternate service provider for these minimal and repetitive tasks.
We also added some "Pepper-centered" answers to questions about the robot's age, name, purpose, feeling, capabilities...

\subsection{Data acquisition}

We placed the robot in the same area as where the reception desk was situated, at the entrance of the university library.
Two large angle cameras were strategically placed in order to grasp the whole scene and especially to understand how users approached the robot before the opening of the interaction. We recorded the audio and video streams from Pepper's tablet. 
As the robot was not programmed to move, it was easier to define a record area.
As regard with personal data protection, posters were placed near the various entrances of the library. After each interaction, a team member obtained signed consent, otherwise the data was deleted.
Eleven recording sessions took place: seven days in March 2022 (17 hours of recording in total) and four days in September 2022 (12 hours of recording in total) as we expected more newcomers at this period of the year in a university library.

\subsection{Naturally-occurring interactions}
\label{naturally}
If we consider the sequential organisation of talk as a mean to secure the appropriate interpretation \textit{for all practical purpose} between two humans, we assume that this minimal understanding procedure between the user and the robot depends on interactional emergent goals brought by the human in front of a seemingly speaking-and-hearing humanoid robot.
Hence, a laboratory setting biases these procedures. For example, users migh pursue the goal of accomplishing a given script, or officially leave the face-to-face configuration having produced reasonably enough turns, overcoming the robot's failures, in order to not disappoint the experimenter....

When we talk about "naturally-occurring interactions" (\cite{mondada2013} for in depth definition and reflection), we point at the fact that interactions were not orchestrated by the researchers. No instructions were given, users were free to interact with the robot and leave whenever they wanted.
One can argue that we intervened in the routines of the library users, which could contradict the "natural" and "unorchestrated" character of the interactions in the ethnographic sense.
However, we disrupted the users habits by the mean of recognizable and acceptable practices: we used a commercial-looking robot that institutions and enterprises use in order to provide basic services while accounting for some technological modernity.
During the recordings, a vast majority of library users thought that it was the library initiative to showcase this robot (despite the posters) until we explained the purpose of its presence.

\subsection{Data format}

The large-angle views and the robot's view recordings were manually synchronized in a video editor. We exported long-format edited multi-scope videos corresponding to each recording sessions (2-3 hours long) that become a temporal reference for time-aligned annotations (\ref{annotation}).
The ELAN\footnote{ELAN (Version 6.4) [Computer software]. (2022). Nijmegen: Max Planck Institute for Psycholinguistics, The Language Archive. Retrieved from \url{https://archive.mpi.nl/tla/elan}} software is used for all the annotation tasks.
The identification and time alignment of the original robot's view recordings have been indicated in a dedicated tier in the annotation software: this way, a script can extract clips corresponding to annotated segments created on other tiers and create different training datasets.

Hence, the data consists of original Pepper's view video files, long-format reference video files, ELAN annotation files, and scripts that extract text (transcriptions and annotations) altogether with original video clips from ELAN annotations.

\subsection{A sequence-oriented labelling scheme for heterogeneous data}
\label{shortclips}
Because we decided to go with naturally-occurring interactions, the corpus is heterogeneous by design. No instructions were given to the users, they did not follow a script with recognizable stages, and the state machine allowed for a large set of combinations (see state machine representation in Appendix \ref{appendix-documentation}).
Before doing time-aligned transcriptions and annotations of human-robot interactions~(\ref{annotation}), we needed to manually define time-aligned shortclips from the long-format video references and characterize them in order to deal with this heterogeneity. 

Thus, we created a labelling scheme and syntax that refer to approximately identified actions that can be part of a sequence. For example, the "greeting1" and "greeting2" tags account for the two parts of a greeting exchange. These tags are entered following their order of appearance in the clips. These are prefixed by letters indicating if the transmitter is the human (h) or the robot (p), and suffixed by the same letters indicating the recipient. This way, we may characterize a whole clip with a string like:

\texttt{hgreeting1p, hquestionp, silence, pgreeting2h}

where we can account for the fact that a human question and a noticeable silence preceded the answered greeting produced by the robot.
This is always the human interpretation that supersedes the interpretation of the robot's action. For example, a same turn produced by the robot "I can provide you information about the library" might be understood as an account for not having responded to a previous request but also as a proposal.
Other sequential and turn-taking features are placed on the same string, like repairs, repeats, or overlaps.
Finally, remarkable and specific resources and phenomena that have been identified in the first analyses of data (internal data sessions) have been added (the fact that the robot might gaze away, when a new eye-contact is established, laughter...).

This way we could identify some first regularities in order to deploy research strategies (with appropriate search strings) and select the more relevant data to segment and transcribe in ELAN, as it is a time-consuming work (around 1 hour per minute of interaction). For a person who is experienced with \textit{CA} and the relevant actions and phenomena identified for the project, such a labelling practice takes 7 minutes on average per minute. A link to the documentation of this labelling scheme is provided in the appendix \ref{appendix-documentation}. It may be used and adapted for any large corpus of heterogeneous interactional data.

\newpage

\section{Annotating interactions}
\label{annotation}
If the labelling system presented above has to deal with heterogeneity, within this section, the annotation system we present has to deal with the temporality and complexity of naturally-occurring interactions. We want to show what it's like to segment transcribed speech segments and ascribe annotated action against a turn-by-turn sequential analysis.
These annotated actions must have their own inter-segment syntax as a mean to account for their temporal and sequential dynamics (normative expectation, abandonment, delaying, repair...).
In the next subsections we show two samples from our corpus. The first sample will allow us to explain the mechanics of such an annotation practice. 
The second sample will show that other resources than talk may receive annotations, especially for the management of turn-taking and byplay participation framework.

\subsection{Sample 1: a multi-threaded sequential infrastructure}
\label{sample1}
When \textit{CA} researchers perform a sequential analysis of a transcribed interaction, they proceed systematically on a turn-by-turn basis~\cite[pp.120-124]{TenHave2007}: they aim at reproducing the online analysis performed by the participants involved.
The idea behind our time-aligned annotation practice is to formalize this analytical process as a mean of standardized annotations temporally embedded.

\begin{figure}[h]
  \centering
  \includegraphics[width=5cm]{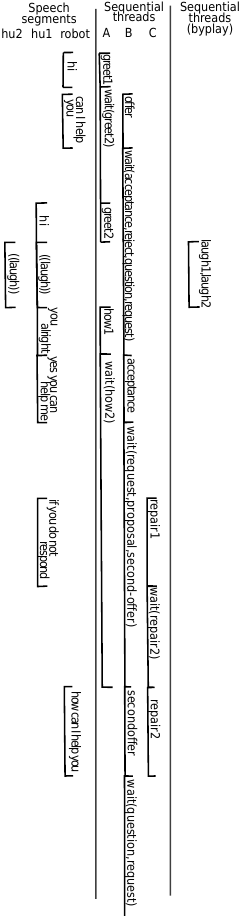}
  \caption{Multi-linear representation of concurrent sequential threads.}
  \label{multi-thread}
  \Description{Multi-linear representation of concurrent sequential threads.}
\end{figure}

We will analyze the piece of data below, extracted from our corpus (see conventions in Appendix \ref{appendix-conventions}). It is the very start of an interaction between two humans (Hum1 and Hum2) and the robot Pepper (Pep) in the university library. Pepper and the humans are in a face-to-face configuration, an eye-contact between Hum1 and Pep just happened. As a matter of readability the turns at talk were directly translated from French:

\begin{verbatim}
1 Pep : hi (.) can I help you?
2        (1.0)
3 Hum1: hi
4        ((hum1 and hum2 laugh))
5 Hum1: you alright? yes you can help me
6        (1.5)
7 Hum1: if you do not respond
8        (2.0)
9 Pep : how can I help you?
\end{verbatim}

Figure \ref{multi-thread} displays a graphical representation of how the annotations a rendered into the time-aligned annotation software ELAN.
We will refer both to the simplified transcript and to the figure. The sequential annotations are results from sequential analysis (\textit{vs.} behavior descriptions or speech transcription only). The vertical axle is temporal and its segmentation is homothetic. 
In the "Speech segments" stream, time segments correspond to utterances that can be isolated as actions (one line per participant).
In the "Sequential threads" stream, sequential labels ("offer", "wait()"...) are annotated with segments aligned with speech segments. The "byplay" sequential threads use the same syntax, but it simply indicates that these actions are not addressed to the robot.
The "threads" are populated depending on free space in the A-B-C order.
\pagebreak

Here follows the turn-by-turn sequential analysis:

Line 1, Pepper produces two recognizable actions packaged into a single turn: a greeting ("hi") and a generic offer "can I help you?". These actions, if recognized as such, project two slots for responses: a second greeting, but also an offer acceptance OR rejection, OR some request/question~\cite[p.101]{kendrickdrew2014}. These are the constraints/resources for the human to produce some next turn. These projections are annotated as "wait(relevant1,relevant2,...)" in the sequential thread A. That means that we give a context for the interpretation of the following silence (in order to discriminate this silence with one produced after a sequence completion).

Line 2, a silence is first noticeable: these long silences are recognized as a HRI~\cite{pelikanbroth2016} specificity and its metrics must be part of the data (in order to discriminate silences interpreted as the absence of response like e.g. line 6).

Line 3, the human produces the projected second greeting. The projection in thread A is stopped/complete. The other projection (in thread B) is maintained as still relevant.

Line 4, the two humans produce laughter in overlap (there are laughter sequences~\cite{glenn_laughter_2003}, like when a first makes a second relevant), in a byplay participation framework\cite{goodwin1990}. These are not addressed to the robot (separated in an other thread) but this activity (vs. a silence) account for delaying the projected actions in thread B.

Line 5, the human produces a non-projected new action (annotated in thread A) in a first segment of her turn ("you alright ?") which projects a new response (projection also annotated adjacently in thread A). Within the same turn, the projected offer acceptance is finally produced (annotated in thread B). Relevantly for a service encounter, an other sequence is now projected either from the robot initiation (a proposal or second offer), or the human may now produce a request.

Line 6, a silence has to be interpreted in the context of two projected types of responses (with two active threads). 
Line 7, the previous silence is designated as a failure. Thus it can be interpreted as a repair initiation (annotated in thread C), the repair being completed depending on the completion of the actions projected on thread A or B OR an account of the abandonment of the repair (like "sorry I didn't understand what you said").
Also, the human withdraws the projected possibility for her to produce a request as the turn-allocation to Pepper is reinforced.
Line 8, an other rather long silence has to be interpreted in the context of three projected types of responses within three threads. 
Line 9, finally, Pepper produces the awaited second offer (designed as a question). The repair is completed, the relevance for a response to "how are you" is abandoned and new actions are projected...

What did we do here ? A lot of the complex dynamics that we analyzed do not appear on the annotated data, nor can be inferred from the sequential threads alone. For example, the dynamics between sequences themselves, informed by the study of service encounters, or silence categorization, nor did we address the fact that the howareyou-sequence had lower relevancy as it was situated in the first part of the human turn line 5.

We used the sequential annotations as a mean to reify actions and projections at the adjacency pair level only. The stacking structures mentioned earlier (\ref{temporally}) may then be approximated, thanks to this multi-threaded approach, by the mean of probabilistic relations between segments and threads. We think that the human "simplistic" inferences about conversational agents (see \ref{discussion}) account for this sequence level of reification, whereas the larger sequential dynamics offer more space for negotiations. 

\subsection{Sample 2: more than text} 
 \label{sample2}
As we mentioned earlier (\ref{temporally}), as the interaction is a continuous process unfolded in time and physical co-presence, other bodily resources gain some relevancy for the meaning making at sequentially relevant slots.
One exemplary case will show how multimodal resources can be analyzed for their contribution to turn-taking management, action ascription, and phenomenons that are specific to HRI such as suspended participation.
In the transcription below, the human verbal response to the robot (a request) starts with "hum" line 12, which is 6,6 seconds after the robot's offer (line 1). Another human is behind:
\small
\begin{verbatim}
01 Pep: Hi (.) can I help you ?
02       (0.4)
03 Hum: (1.0) ((starts torquing away))
04 Hum: ((laugh while orienting back towards pepper))
05      (0.9)
06 Hum: ((inhale demonstrably))
07      (0.3)
08       (0.2) ((starts torquing away))
09 Hum: but hum what do I ask?
10 Hum: ((orients back towards pepper))
11      (1.6)
12 Hum: hu::::m
13      (0.5)
14 Hum: I'm looking for a biology book.
\end{verbatim}
\normalsize
If we consider this transcribed data with a verbocentric approach, we have to wait for the (rather recurrent) byplay question used as a delaying device as a cue (after 3,6 seconds of silence) in order to account for the fact that the human will try to respond to the robot. However, this turn (and the laughter l.4 as in \ref{sample1}) that the human addresses to another participant is recorder with a lower voice intensity, which could lead to the recognition of speech only line 12, after 6,6 seconds of silence which is rather long.

We may now consider all the bodily conducts and vocalizations produced in this interaction. 
We can see that it is only 0.4 seconds after Pepper's offer that the human turn around towards her fellow. A relevant bodily cue here is the fact that she accomplishes a \emph{body torque} with the head directed towards the human behind while her legs are still oriented towards Pepper.
This resource has been identified~\cite{Schegloff1998} as indicating an instability that project a short end: the head is oriented towards a temporary interaction goal (first a laugh l.4 then a question l.9) while the lower body part indicate the main interactional focus: interacting with Pepper.
It results that the whole body is in a recognizable torque position with the shoulders and torso oriented towards nothing in particular: they appear sideways from Pepper's view. Moreover, the displayed inhalation in front of the robot (l.6) is also a cue of turn pre-beginning~\cite{mondada_lenonciation_2016}.
In other words, we can rely on these cues in order to recognize the fact that he human is actually preparing a response l.3, which is only 0,4secs after the robot's offer, and then have an additional cue l.6, after 2,8 seconds (vs. 3,6 or even 6,6 seconds of silence that could be interpreted as a disengagement).

This sample shows that a \textit{torque}, if recognized, can contextually provide cues about what comes next (response relevance maintained), turn management (delay), participation (byplay). Being relevant for byplay sequences, its recognition could also inform us about practices where humans assess the robot's behaviour/response after sequence completion, as it is frequent in our corpus.

\section{Discussion: human's perspective and statistical perspective}
\label{discussion}
If we consider the purposefulness of the use case, interactions might appear quite specific.
But as humans appear to draw on generic resources for making sense of their first encounter with a robot, data show that they invoke basic sequences of action (offers, requests, proposals, questions, instructions, greetings, closings) as a way to secure their participation.
This "basicness" was also a feature when the software was designed.

Our corpus suggests that humans already infer basic features, probably from the use of other conversational systems and devices~\cite{Porcheron2018}).
They use intonation emphasis on what appears to be the most relevant keyword to recognize for them (see also~\cite{avgustisetal2021}).
They allow longer silences between turns (\ref{sample1} but also in~\cite{pelikanbroth2016}).
They may even suspend the participation framework with Pepper at every moment, by the mean of torques such as in \ref{sample2}.
They also perform less actions per turn, giving back the turn-at-talk to the robot, as in \ref{sample1} where the offer acceptance is not immediately followed by the request as compared with human-human interaction.
In other words, natural HRI show that acting in a simulacrum of conversation~\cite{button1990} raise recognized and established practices such as those exemplified above, which contributes to a better ecology between the human and the conversational agent.

By reifying sequences of actions, we do not aim at replicating an interactional competence, especially because humans do not use or learn statistically such sequential features (for e.g. see~\cite{Filipi2018} for child acquisition).
Our goal is to accompany the rational practical work accomplished by the human that is aware of being talking with a machine (see~\cite{alac_when_2011} for this \textit{ethnomethodological} perspective).

One of the limits of the annotation system we proposed is the quantity of data that can be annotated, as qualified CA researchers must perform it. Once this qualitative-oriented annotation system is stabilized, we will also assess inter-rater reliability.

Actual natural language understanding models are able to learn predictive word models and to recognize intents, even with little data (thanks to few-shot learning)~\cite{xia2022natural,brown2020}. 
As a perspective our work may improve these AI conversational systems by coupling the intentions detected by the system (learned thanks to our annotated data) with the turn taking sequences we began to identify, to make the conversation more natural. While some errors of the system can be tolerated and corrected by humans that adapt their behavior to an artificial entity (as observed in our data), we may even study how the robot can improve in the wild by interacting with human users and progressively refine the detected intention and sequences.

\begin{acks}
The authors are grateful to the ASLAN project (ANR-10-LABX-0081) of the Université de Lyon, for its financial support within the French program "Investments for the Future" operated by the National Research Agency (ANR).
\end{acks}
\bibliographystyle{ACM-Reference-Format}
\bibliography{bibliography}

\appendix
\section{Conventions}
\label{appendix-conventions}
Transcript conventions, drastically simplified from ICOR/Jefferson:
\begin{itemize}
    \item \texttt{(.)} perceptible silence <200ms
    \item \texttt{(1.0)} mesured length of a silence >200ms in seconds
    \item \texttt{:} prolongation of the immediately prior sound (impressionistic representation with additional colons)
    \item \texttt{?} a raising intonation (not a question mark per se, as raising intonations might appear at the end of other types of utterances)   
    \item \texttt{((event))} events or conducts that could not be transcribed
\end{itemize}

\section{Project documentation}
\label{appendix-documentation}

\begin{itemize}
    \item The state machine graphical representation of the dialogue system ad hoc version can be found here: \url{https://page.hn/shhg8j}
    \item The documentation of the labelling system for annotating shortclips can be found here: \url{https://page.hn/0fkplh}
\end{itemize}

\end{document}